\pdfoutput=1

\documentclass[11pt, dvipsnames, table]{article}

\usepackage{EACL2023}

\usepackage{times}
\usepackage{latexsym}

\usepackage[T1]{fontenc}

\usepackage[utf8]{inputenc}

\usepackage{microtype}
\usepackage{graphicx}
\usepackage{amsmath}
\usepackage{bbm}
\usepackage{amssymb}
\usepackage{amsfonts}
\usepackage{booktabs}
\usepackage{lipsum}
\usepackage{multicol}
\usepackage{makecell}
\usepackage{rotating}
\usepackage{multirow}
\usepackage{misra}
\usepackage{tikz}
\usepackage{caption}
\usepackage{subcaption}
\usepackage{xcolor}
\usepackage{enumitem}
\usepackage{graphicx}
\usepackage{linguex}
\usepackage{enumitem}
\usepackage{highlight}
\usepackage{wasysym}
\usepackage{soul}
\usepackage{ragged2e}
\usepackage{array}
\usepackage{framed}
\usepackage{inconsolata}
\usepackage{pifont}
\usepackage[most]{tcolorbox}
\usepackage[frozencache,cachedir=.]{minted}
\definecolor{yell}{HTML}{FBEEAC}

\newcommand{\cmark}{\ding{51}}%
\newcommand{\xmark}{\ding{55}}%

\newcommand{\comps}{\textsc{comps}}
\newcommand{\minicomps}{mini\textsc{comps}}

\newcommand{\minicompswugs}{mini\textsc{comps-wugs}}
\newcommand{\minicompswugsdist}{mini\textsc{comps-wugs-dist}}
\newcommand{\compsbase}{\textsc{comps-base}}
\newcommand{\compswugs}{\textsc{comps-wugs}}
\newcommand{\compswugsdist}{\textsc{comps-wugs-dist}}

\newcommand{\plms}{PLMs}

\definecolor{pblu}{HTML}{8AC4D0}
\definecolor{pyello}{HTML}{F4D160}

\newcommand{\plotblue}{\textcolor{pblu}{\textbf{blue}}}
\newcommand{\plotyellow}{\textcolor{pyello}{\textbf{yellow}}}

\definecolor{orang}{HTML}{d95f02}
\definecolor{purpl}{HTML}{7570b3}

\newcommand{\befor}{\textcolor{orang}{\textbf{before}}}
\newcommand{\inbetween}{\textcolor{purpl}{\textbf{in-between}}}

\newcommand{\nproperties}{3,645}
\newcommand{\nminpairs}{49,280}

\newcommand{\blank}{$\rule{0.6cm}{0.15mm}$}

\definecolor{lightmauve}{rgb}{0.86, 0.82, 1.0}
\definecolor{royalazure}{rgb}{0.0, 0.22, 0.66}

\title{\textsc{comps}: Conceptual Minimal Pair Sentences for testing Robust Property Knowledge and its Inheritance in Pre-trained Language Models}

\author{Kanishka Misra\\
  Purdue University \\
  \texttt{kmisra@purdue.edu}\\
  \And
  Julia Rayz\\
  Purdue University \\
  \texttt{jtaylor1@purdue.edu}\\
  \And
  Allyson Ettinger\\
  University of Chicago \\
  \texttt{aettinger@uchicago.edu}
 }

\begin{document}

\maketitle

\begin{abstract}
A characteristic feature of human semantic cognition is its ability to not only store and retrieve the properties of concepts observed through experience, but to also facilitate the inheritance of properties (\textit{can breathe}) from superordinate concepts (\textsc{animal}) to their subordinates (\textsc{dog})---i.e. demonstrate \textit{property inheritance}. 
In this paper, we present \comps{}, a collection of English minimal pair sentences that jointly tests pre-trained language models (PLMs) on their ability to attribute properties to concepts and their ability to demonstrate property inheritance behavior. 
Analyses of 22 different PLMs on \comps{} reveal that they can easily distinguish between concepts on the basis of a property when they are trivially different, but find it relatively difficult when concepts are related on the basis of nuanced knowledge representations. Furthermore, we find that PLMs can show behaviors suggesting successful property inheritance in simple contexts, but fail in the presence of distracting information, which decreases the performance of many models sometimes even below chance.
This lack of robustness in demonstrating simple reasoning raises important questions about PLMs' capacity to make correct inferences even when they appear to possess the prerequisite knowledge.
\end{abstract}

\section{Introduction}
\label{sec:intro}
The ability to learn, update and deploy one’s knowledge about concepts (\textsc{robin}, \textsc{chair}) and their properties (\textit{can fly}, \textit{can be sat on}), observed during everyday experience is fundamental to human semantic cognition \citep{murphy2004big,  rogers2004semantic, rips2012concepts}. 
{Knowledge of a concept's properties, combined with the ability to infer the \texttt{IsA} relation \citep{sloman1998categorical, murphy2003semantic} leads to an important behavior known as} \textit{property inheritance} \citep{quillian1967word, smith1978theories, murphy2004big}, where subordinates of a concept inherit its properties.
For instance, one is likely to infer that an entity called \emph{luna} can meow, has a tail, is a mammal, etc., even if \emph{all} they know is that it is a cat. 
The close connection between a word's meaning and its conceptual representation makes these abilities crucial to language understanding \citep{murphy2004big, lake2021word}, making it critical for computational models of language processing to also exhibit behavior consistent with these capacities. 
Indeed, modern pre-trained language models \citep[\plms{};][etc.]{devlin-etal-2019-bert, brown2020language} have made impressive empirical strides in eliciting general knowledge about real world concepts and entities \citep[\textit{i.a.}]{petroni2019language, weir2020probing}, as well as in demonstrating isomorphism with real world abstractions like direction and color \citep{abdou-etal-2021-language, patel2022mapping}, often times without even having been explicitly trained to do so.
At the same time, their ability to robustly demonstrate such capacities has recently been called to question, owing to failures due to reporting bias \citep{gordon2013reporting, shwartz-choi-2020-neural}, lack of consistency \citep{elazar-etal-2021-measuring, ravichander-etal-2020-systematicity}, and sensitivity to lexical cues \citep{kassner-schutze-2020-negated, misra-etal-2020-exploring, pandia-ettinger-2021-sorting}.

In this work, we cast further light on \plms{}' ability to robustly demonstrate knowledge about concepts and their properties. 
To this end, we introduce Conceptual Minimal Pair Sentences (\comps{}), a collection of English minimal pair sentences, where each pair attributes a property (\textit{can fly}) to two noun concepts: one which actually possesses the property (\textsc{robin}), and one which does not (\textsc{penguin}).
Following standard practice in the minimal pairs evaluation paradigm \citep[etc.]{warstadt-etal-2020-blimp-benchmark}, we test whether \plms{} prefer sentence stimuli expressing correct property knowledge over those expressing incorrect ones.
\comps{} can be decomposed into three subsets, each containing stimuli that progressively isolate deeper understanding of the task of attributing properties to concepts, by adding controls for more superficial heuristics.
Our first subset---\compsbase{}---measures the extent to which \plms{} attribute properties to the right concepts, while varying the similarity of the positive (\textsc{robin}) and the negative concepts (\textsc{penguin} [high] vs. \textsc{table} [low]). 
This controls for {the possibility that models are relying on} coarse-grained concept distinctions. {For instance, in this setup} a model should prefer \cref{ex:acc} over both versions of \cref{ex:unacc}.
\ex.\label{ex:compsbase}
    \a.\label{ex:acc} A \textbf{robin} can fly.
    \b.\label{ex:unacc} *A (\textbf{penguin/table}) can fly.
    
Next, drawing on the phenomenon of property inheritance, the \compswugs{} set introduces a novel concept, \textsc{wug}, expressed as the subordinate of the positive and negative concepts from a subset of the \compsbase{} set, and tests the extent to which \plms{} successfully attribute it the given property when it is associated with the positive concept. 
{This increases the complexity of the reasoning task, as well as the distance between the associated concept (\textsc{robin}) and property (\textit{can fly}). These manipulations help to control for memorization of the literal phrases being tested, forcing models to judge properties for a novel concept that inherits the property from a known concept.}
In this task, given that a model successfully prefers \cref{ex:acc} over \cref{ex:unacc}, it should also prefer \cref{ex:wugsacc} over \cref{ex:wugsunacc}:
\ex.\label{ex:compswugs}
\small
    \a.\label{ex:wugsacc} A wug is a \textbf{robin}. Therefore, a wug can fly.
    \b.\label{ex:wugsunacc} *A wug is a \textbf{penguin}. Therefore, a wug can fly.

The final subset---\compswugsdist{}, combines the aforementioned controls by using negative concepts as distracting content and inserting them into the \compswugs{} stimuli. Specifically, we transform the stimuli of \compswugs{} by creating two subordinates for every minimal pair; one for the positive concept (\textsc{robin}, subordinate: \textsc{wug}) and the other for the negative concept (\textsc{penguin}, subordinate: \textsc{dax}), which acts as a distractor. 
This way, {we control for the possibility that models may be relying on simple word associations between content words---of which there are only two in the prior tests---by introducing additional, irrelevant but contentful words into the context.}
Here, we consider models to be correct if they prefer \cref{ex:wugsdistacc} over \cref{ex:wugsdistunacc}, given that they prefer \cref{ex:acc} over \cref{ex:unacc}:

\ex.\label{ex:compswugsdist}
\small
    \a.\label{ex:wugsdistacc} A \textbf{wug} is a robin. A \textbf{dax} is a penguin. Therefore, a \textbf{wug} can fly.
    \b.\label{ex:wugsdistunacc} *A \textbf{wug} is a robin. A \textbf{dax} is a penguin. Therefore, a \textbf{dax} can fly.

{Together, the three sets of stimuli tease apart more superficial predictive behaviors, such as contextual word associations, from more robust reasoning behaviors based on understanding of concept properties. While we can expect superficial predictive strategies to be brittle in the face of shallow perturbations and irrelevant distractions, robust property knowledge and reasoning behaviors should not}. 

We use \comps{} to analyze robust property knowledge and its inheritance in 22 different \plms{}, ranging from small masked language models to billion-parameter autoregressive language models.
In our experiments with \compsbase{}, we find \plms{} to demonstrate strong performance in attributing properties to the correct concepts in our minimal pairs. 
However, we observe this strong performance largely when the concepts in the minimal pairs are trivially different (e.g., \textsc{lion} and \textsc{tea} for the property \textit{is a mammal}).
When the concept pairs are similar (on the basis of different knowledge representations), we find models' performance to degrade substantially, by as much as 25 points.
We observe a similar trend in our analyses on \compswugs{}---models first appear to show desirable behavior, potentially indicating proficiency in the more complex property inheritance reasoning. 
However, their overall performance declines drastically when investigated in the presence of distractors (i.e., on \compswugsdist{}).
This failure is particularly pronounced in larger autoregressive \plms{}, whose performance in fact drops below chance in cases where distracting information is proximal to the queried property, indicating the presence of a proximity effect.
Together, our findings highlight brittleness of \plms{} with conceptual knowledge and reasoning, as evidenced by failures in the face of simple controls.
We make our code and data available at: \url{https://github.com/kanishkamisra/comps}.

\section{Conceptual Minimal Pair Sentences (\comps{})}
\label{sec:comps}
\subsection{Connections to prior work}
Prior work in exploring property knowledge in \plms{} has adopted two different paradigms: one which uses probing classifiers to test if the applicability of a property can be decoded from the representations of LMs \citep{forbes2019neural, da-kasai-2019-cracking, derby-etal-2021-representation}; and the other which uses cloze-testing, in which LMs are tasked to fill in the blank in prompts that describe specific properties/factual knowledge about the world \citep{petroni2019language, weir2020probing}. We argue that both approaches---though insightful---have key limitations for evaluating property knowledge, and that minimal pair testing overcomes these limitations to a beneficial extent.

Apart from ongoing debates surrounding the validity of probing classifiers \citep[see][]{hewitt2019designing, ravichander-etal-2021-probing, belinkov2022probing}, the probing setup does not allow the testing of property knowledge in a precise manner. Specifically, several properties are often perfectly correlated in datasets such as the one we use here (see \cref{sec:gtruth}). For example, the property of being an animal and being able to breathe and grow, etc., are all perfectly correlated with one another. Even if the model’s true knowledge of these properties is highly variable, probing its representations for them would yield the exact same result, leading to conclusions that overestimate the model's capacity for some properties, while underestimating for others.
Evaluation using minimal pair sentences overcomes this limitation by allowing us to explicitly represent the properties of interest in language form, thereby allowing precise testing of property knowledge.

Similarly, standard cloze-testing of \plms{} \citep{petroni2019language, weir2020probing, jiang2021can} also faces multiple limitations. First, it does not allow for testing of multi-word expressions, as by definition, it involves prediction of a single word/token. Second, it does not yield faithful conclusions about one-to-many or many-to-many relations: e.g. the cloze prompts ``Ravens can \blank{}.” and ``\blank{} can fly.” do not have a single correct answer. This makes our conclusions about models’ knowledge contingent on choice of one correct completion over the other.
The minimal pair evaluation paradigm overcomes these issues by generalizing the cloze-testing method to multi-word expressions---by focusing on entire sentences---and at the same time, pairing every prompt with a negative instance. 
This allows for a straightforward way to assess correctness: the choice between multiple correct completions is transformed into one between correct and incorrect, at the cost of having several different instances (pairs) for testing knowledge of the same property.
Additionally, the minimal pairs paradigm allows us also to shed light on how the nature of negative samples affects model behavior, which has been missing in approaches using probing and cloze-testing. 
The usage of minimal pairs is a well-established practice in the literature, having been widely used in works that analyze syntactic knowledge of LMs \citep{marvin2018targeted, futrell2019neural, warstadt-etal-2020-blimp-benchmark}. 
We complement this growing literature by introducing minimal-pair testing to the study of conceptual knowledge in \plms{}.

Our property inheritance analyses closely relate to the `Leap-of-Thought' (LoT) framework of \citet{talmor2020leap}. In particular, LoT holds the taxonomic relations between concepts \textit{implicit} and tests whether models can abstract over them to make property inferences---e.g., testing the extent to which models assign \textit{Whales have bellybuttons} the `True' label, given that \textit{Mammals have bellybuttons} (with the implicit knowledge here being \textit{Whales are mammals}).
With \compswugs{} (and \compswugsdist{}), we instead explicitly provide the relevant taxonomic knowledge {in the context} and target whether \plms{} can behave consistently with knowledge they have already {demonstrated} (in the base case, \compsbase{}) {and attribute the property in question to the correct subordinate concept}. This also relates to recent work that measures consistency of \plms{}' word prediction capacities in eliciting factual knowledge \citep{elazar-etal-2021-measuring,ravichander-etal-2020-systematicity}.

\subsection{Ground-truth Property Knowledge data}
\label{sec:gtruth}

For our ground-truth property knowledge resource, we use a subset of the CSLB property norms collected by \citet{devereux2014centre}, which was further extended by \citet{misra2022property}. The original dataset was constructed by asking 123 human participants 
to generate properties for 638 everyday concepts.
Contemporary work has used this dataset by taking as positive instances all concepts for which a property was generated, while taking the rest as negative instances \citep[][etc.]{lucy-gauthier-2017-distributional, da-kasai-2019-cracking} for each property.
While this dataset has been popularly used in related literature, \citet{misra2022property} recently discovered striking gaps in coverage among the properties included in the dataset.\footnote{See also \citet{sommerauer-fokkens-2018-firearms} and \citet{sommerauer2022diagnosing}, who also discuss this limitation.} For example, the property \textit{can breathe} was only generated for 6 out of 152 animal concepts, despite being applicable for all of them---as a result, contemporary work can be expected to have wrongfully penalized models that attributed this property to animals that could indeed breathe, and similarly for other properties.
To remedy this issue, \citet{misra2022property} manually extended CSLB's coverage for 521 concepts and \nproperties{} properties. We refer to this extended CSLB dataset as XCSLB, and we use it as our source for ground-truth property knowledge.

\subsection{Choosing negative samples}
\label{sec:compsneg}

We rely on a diverse set of knowledge representation sources to construct negative samples for \comps{}. Each source has a unique representational structure which gives rise to different pairwise similarity metrics, on the basis of which we pick out negative samples for each property:
\paragraph{Taxonomy} We consider a hierarchical organization of our concepts, by taking a subset of WordNet \citep{miller1995wordnet} consisting of our 521 concepts. We use the \texttt{wup} similarity \citep{wu-palmer-1994-verb} as our choice of taxonomic similarity.
\paragraph{Property Norms} We use the XCSLB dataset and organize it as a matrix whose rows indicate concepts and columns indicate properties that are either present (indicated as 1) or absent (indicated as 0) for each concept. As our similarity measure, we consider the jaccard similarity between the row vectors of concepts. This reflects the overlap in properties between concepts, and is prevalent in studies utilizing conceptual similarity in cognitive science \citep[][etc.]{tversky1977features, sloman1993feature}.
\paragraph{Co-occurrence} We use the co-occurrence between concept words as an unstructured knowledge representation. For quantifying similarity, we use the cosine similarity of the GloVe vectors \citep{pennington2014glove} of our concept words.

\paragraph{Sampling Strategy}  Each property ($p_i$) in our dataset splits the set of concepts into two: a set of concepts that possess the property ($Q_{p_i}$), and a set of concepts that do not ($\neg Q_{p_i}$).
We sample $\min(|Q_{p_i}|, 10)$---i.e., at most 10---concepts from $Q_{p_i}$ and take them to be our positive set. Then for each concept in the positive set, we sample from $\neg Q_{p_i}$ the concept that is most similar (depending on the source) to the positive concept and take it as a negative concept for the property.
We additionally include a negative concept that is randomly sampled from $\neg Q_{p_i}$, leaving out the concepts sampled on the basis of the three previously described knowledge sources.
Examples of the four types of negative samples for the concept \textsc{zebra} and the property \textit{has striped patterns} are shown in \Cref{tab:example}. 

\begin{table}[!t]
\centering
\resizebox{0.8\columnwidth}{!}{%
\begin{tabular}{@{}lcc@{}}
\toprule
\textbf{Knowledge Rep.} & \textbf{Negative Concept} & \textbf{Similarity} \\ \midrule
Taxonomy & \textsc{horse} & 0.88 \\
Property Norms & \textsc{deer} & 0.63 \\
Co-occurrence & \textsc{giraffe} & 0.75 \\
Random & \textsc{bat} & - \\ \bottomrule
\end{tabular}%
}
\caption{Negatively sampled concepts selected on the basis of various knowledge representational mechanisms, where the property is \textit{has striped patterns}, and the positive concept is \textsc{zebra}.}
\label{tab:example}
\vspace{-1.2em}
\end{table}

\begin{figure*}[t]
    \centering
    \begin{subfigure}[b]{0.47\textwidth}
    \small
        \begin{quote}
            \textbf{Property:} \textit{can fly}\\
            \textbf{Positive:} \textsc{robin}\\
            \textbf{Negative:} \textsc{penguin}\\
            \textbf{Subordinate:} \textsc{wug}\\
             \textbf{\compsbase{}:} A (\textbf{robin/penguin}) can fly.\\
            \textbf{\compswugs{}:} A wug is a (\textbf{robin/penguin}). Therefore, a wug can fly.
        \end{quote}
        \caption{Instances of \compsbase{} and \compswugs{}.}
        \label{fig:basewugsstim}
    \end{subfigure}
    \hfill
    \begin{subfigure}[b]{0.52\textwidth}
        \centering
         \includegraphics[width=0.9\columnwidth]{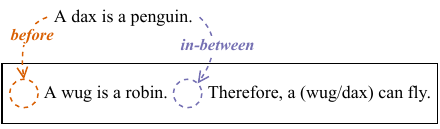}
        \caption{Distraction scheme for stimuli in \compswugsdist{}, where the distractor is inserted either \textbf{\textcolor{orang}{before}} or \textbf{\textcolor{purpl}{in between}} each \compswugs{} stimulus.}
        \label{fig:wugsdiststim}
    \end{subfigure}
    \caption{Examples of materials used in our experiments. In this example, \textsc{robin} is the positive concept.}
    \label{fig:compsstim}
    \vspace{-1em}
\end{figure*}

\subsection{Minimal Pair Construction}
\label{sec:stimuli}
Following our negative sample generation process, we end up with total of \nminpairs{} pairs of positive and negative concepts
that span across \nproperties{} properties (14 pairs per property, on average). Every property is associated with a property phrase---a verb phrase which expresses the property in English, as provided in XCSLB. Using these materials, we construct our three datasets of minimal pair sentence stimuli, examples of which are shown in \Cref{fig:compsstim}.

\paragraph{\compsbase{}} The \compsbase{} dataset contains minimal pair sentences that follow the template: ``\texttt{[DET] [CONCEPT] [property-phrase]}.'' where \texttt{[DET]} is an optional determiner, and \texttt{[CONCEPT]} is the noun concept. 
Applying this template to our generated pairs results in \nminpairs{} instances. See \Cref{fig:basewugsstim} for an example.
\paragraph{\compswugs{}} We test property inheritance in \plms{} using only the animal kingdom subset of \compsbase{} (152 concepts, 944 properties, and 13,888 pairs), keeping the same negative samples. 
We convert the original minimal pair sentences in \compsbase{}, in which the positive concept is an animal, into pairs of two-sentence stimuli by first introducing a new concept (\textsc{wug}) to be the subordinate of the concepts in the original minimal pair. We then express its property inheritance in a separate sentence. Our two sentence stimuli follow the template: ``A wug is a \texttt{[CONCEPT]}. Therefore, a wug \texttt{[property-phrase]}.'' {Although we use \textit{wug} as our running example for the subordinate concept, we use four different nonsense words \{\textit{wug}, \textit{dax}, \textit{blicket}, \textit{fep}\} equal numbers of times, to avoid making spurious conclusions based on a single nonsense word}.\footnote{As we describe in \cref{sec:experiments}, we also tried a different set of nonce words, {to address concerns about possible impacts of using nonce words from existing literature (e.g., \emph{wug})}.} 
Introducing an intervening novel concept allows us to robustly control for simple word-level associations between concepts and properties that models might have picked up during training.
\Cref{fig:basewugsstim} shows an example.

\paragraph{\compswugsdist{}} To add distracting information, we follow \citet{pandia-ettinger-2021-sorting} and convert the \compswugs{} stimuli by associating a different subordinate concept (\textsc{dax}) with the negative concept (\texttt{[NEG-CONCEPT]}), and inserting it \befor{} or \inbetween{} the sentence containing the positive concept and its subordinate, separately. This results in two subsets (\befor{} and \inbetween{}) of three-sentence minimal pair stimuli, which differ in the subordinate to which the property is attributed. We use the following template to create our stimuli: ``A \textbf{wug} is a \texttt{[CONCEPT]}. A \textbf{dax} is a \texttt{[NEG-CONCEPT]}. Therefore, a (\textbf{wug}/\textbf{dax}) \texttt{[property-phrase]}.'' That is, we have stimuli that resemble \compswugs{} but instead deal with a pair of competing subordinate concepts in context.\footnote{{We again choose from our list of  four nonsense words (\textit{wug}, \textit{dax}, \textit{blicket}, and \textit{fep}), which amounts to 12 unique ordered pairs, after accounting for counterbalancing.\label{foot:counterbalanced}}} 
See \Cref{fig:wugsdiststim} for an example.

\section{Methodology}
\subsection{Models Investigated}
We investigate property knowledge and property inheritance capacities of 22 different \plms{}, belonging to six different families. We evaluate four widely used masked language modeling (MLM) families: (1) ALBERT \citep{lan2019albert}, (2) BERT \citep{devlin-etal-2019-bert}, (3) ELECTRA \citep{clark2020electra}, and (4) RoBERTa \citep{liu2019roberta}; as well as two auto-regressive language modeling families: (1) GPT2 \citep{radford2019language}, and (2) the GPT-Neo \citep{gpt-neo} and GPT-J models \citep{gpt-j} from EleutherAI. 
We also use distilled versions of BERT-base, RoBERTa-base, and GPT2, trained using the method described by \citet{sanh2019distilbert}.
We list each model's parameters, vocabulary size, and training corpora in \Cref{tab:modelmeta} (\Cref{sec:modelmeta}).

\begin{figure*}[t!]
    \centering
    \includegraphics[width=\textwidth]{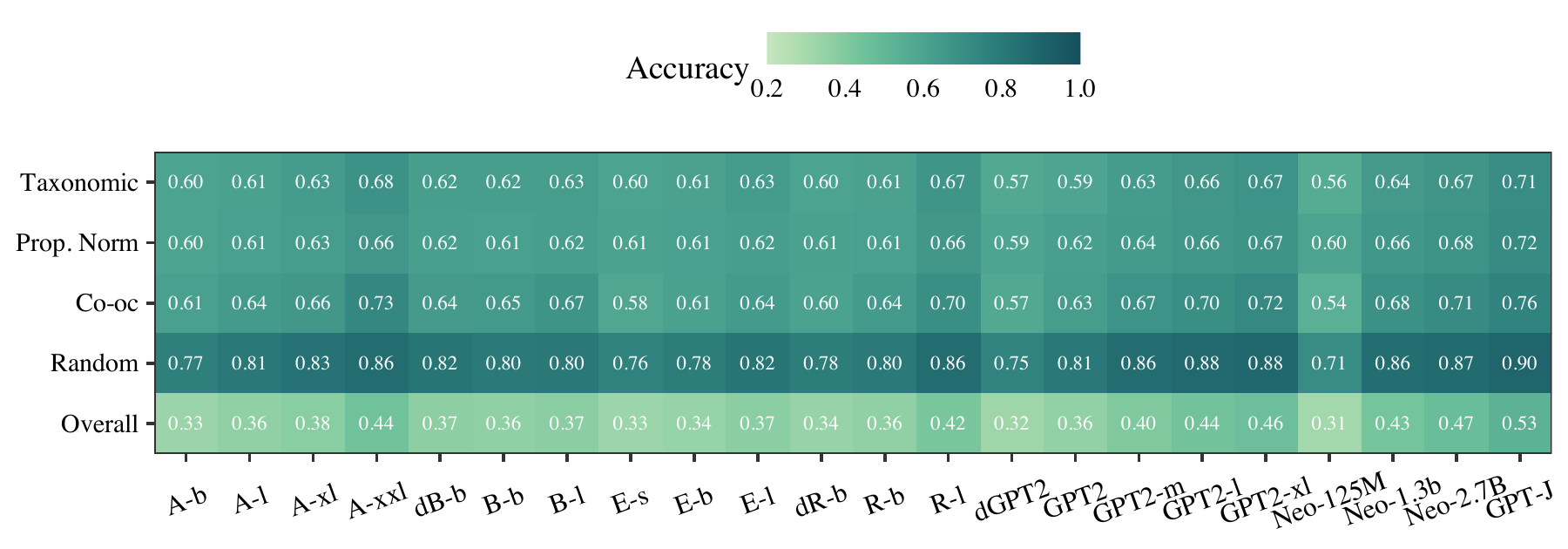}
    \caption{Accuracies of \plms{} on \compsbase{} under various negative sampling schemes. Chance performance for all rows is 50\%, except for `Overall,' where it is 6.25\%. Refer to \Cref{tab:modelmeta} for unabbreviated model names.}
    \label{fig:compsall}
    \vspace{-1em}
\end{figure*}

\subsection{Measuring Performance}
\label{sec:performance}
To evaluate models on \comps{}, we compare their log-probabilities for the property phrase---conditioned on contexts (to the left) containing the positive and negative noun concepts. 
That is, we hold the property phrase constant, and compare across minimally differing conditions to evaluate the probability with which a property is attributed to each concept. 
For example, we score stimuli in \compsbase{}, e.g., ``\textit{A dog can bark.}'' as:
\begin{align*}
    \log p (\text{can bark.}& \mid \text{A dog}),
\end{align*}
its corresponding stimulus in \compswugs{}, ``\textit{A wug is a dog. Therefore, a wug can bark.}'' as:
\begin{align*}
    \small \log p (\text{can bark.} \mid \text{A wug is a dog. Therefore, a wug}),
\end{align*}
and similarly---assuming \textsc{cat} as the negative concept---the corresponding stimuli in our \compswugsdist{} subset, ``\textit{A wug is a dog. A dax is a cat. Therefore, a wug can bark.}'' as:\footnote{Here we show an example where the distractor is added \inbetween{} the context specifying the positive concept, and the queried property knowledge.}
\begin{align*}
    \small \log p (\text{can bark.} \mid &\small\text{ A wug is a dog. A dax is a cat. There-}\\
     &\small\text{ fore, a wug}).
\end{align*}
This approach to eliciting conditional LM judgments is equivalent to the ``scoring by premise'' method \citep{holtzman-etal-2021-surface}, which has been shown to result in stable comparisons across items. 
Additionally, this also takes into account the potential noise due to frequency effects or tokenization differences \citep{misra2021typicality}. 
Estimating these conditional log-probabilities using auto-regressive \plms{} can be directly computed in a left-to-right manner. For MLMs, we use their conditional pseudo-loglikelihoods \citep{salazar-etal-2020-masked} as a proxy for conditional log-probabilities.

Based on this simple method of eliciting relative acceptability measures from \plms{}, we evaluate a model's accuracy on all \comps{} stimuli as the percentage of times its log-probability for a property is greater when conditioned on the context that attributes the property to the positive---as opposed to the negative---concept. Since all cases are forced-choice tasks between two instances, chance performance is set to 50\%.
\Cref{tab:examplescores} (\Cref{sec:appstimuli}) shows examples of all \comps{} stimuli and GPT-J's conditional log-probabilities for them.

\section{Experiments and Analyses}
\label{sec:experiments}
\subsection{Base property knowledge of PLMs and their sensitivity to similarity effects}
\label{sec:base}
We begin by evaluating the 22 \plms{} on \compsbase{}.
Here we focus on the extent to which models robustly associate properties to the correct concepts across stimuli with varying kinds of similarity between the positive and negative concepts.
We report accuracies of the 22 \plms{} on \compsbase{} across the four different negative sampling schemes that we specified in \Cref{sec:compsneg}. 
We additionally report a more stringent accuracy measure that we refer to as `Overall accuracy,' which is calculated for every property and its positive concept, as the percentage of times a model correctly attributes the property to the positive concept in \textbf{all four types of negative sampling schemes}. Chance performance for only the `Overall' case is then 6.25\% ($\text{0.5}^4 \times \text{100}$). \Cref{fig:compsall} shows these results.

From \Cref{fig:compsall}, we see that models strongly distinguish between positive and negative concepts in cases where they are dramatically different---i.e., where negative concepts were sampled randomly (e.g., \textsc{bear} [positive] vs \textsc{bottle} [negative] for the property \textit{can breathe}). 
However performance drops substantially when there are subtler differences between the two concepts---e.g, the concepts \textsc{walrus} (positive) and \textsc{shark} (negative) for the property \textit{is a mammal}.
For instance, the best performing model in any similarity-based negative sampling scheme (GPT-J, 76\%, `Co-oc') only slightly outperforms the worst model in the random negative sampling scheme (Neo-125M, 71\%).
The performance of \plms{} is not substantially different across the three similarity-based negative sampling schemes, 
suggesting that the dynamics of model sensitivity in attributing properties to concepts are largely harmonized across various types of similarities.
As a result of models' insensitivity in presence of similar negative concepts, the overall accuracies are very modest in value, with the overall accuracy of the best performing model (GPT-J) being only 53\%. 
This overall performance is, however, significantly above chance (6.25\%).
We discuss additional findings, such as performance by property type and model size, in \Cref{sec:extra}, since they are incidental to the main conclusions of this analysis.

\subsection{Property inheritance in \plms{}}
Having established the base property knowledge of \plms{}, we now investigate the extent to which they can show behavior that is consistent with reasoning required to handle property inheritance.
We first investigate their performance on \compswugs{}, created using the subset of \compsbase{} containing only animal concepts (see \Cref{sec:stimuli} for stimulus construction). \Cref{tab:overall} shows average accuracies obtained by \plms{} on our property inheritance stimuli, and compares them to average accuracies on \compsbase{}---aggregating across all negative sampling schemes. 
Recall that the stimuli in \compswugs{} present a more challenging {property attribution} task than in \compsbase{}, by not only controlling for coarse-grained similarity effects, but also introducing an intervening novel concept that is expected to inherit the properties of the positive concept.
By measuring attribution of properties more indirectly, these stimuli {increase the complexity of the reasoning and control for memorization of the literal phrase initially tested with \compsbase{}.}

\begin{table}[!t]
\centering
\resizebox{0.7\columnwidth}{!}{%
\begin{tabular}{@{}lcc@{}}
\toprule
\textbf{\textsc{comps} subset} & \textbf{Size} &\textbf{Acc.}\\ \midrule
\textsc{base} & 49.3K & $\text{68.4}_{\text{1.7}}$ \\
\textsc{base} (\textit{animal kingdom only}) & 13.8K & $\text{67.1}_{\text{2.0}}$ \\ \midrule
\textsc{wugs} & 13.8K & $\text{68.9}_{\text{2.3}}$ \\
\textsc{wugs-dist} (\befor) & 13.8K & $\text{59.2}_{\text{3.9}}$ \\
\textsc{wugs-dist} (\inbetween) &  13.8K & $\text{47.2}_{\text{4.5}}$\\ \bottomrule
\end{tabular}%
}
\caption{Average accuracy (and standard error of the mean) of \plms{} ($N=\text{22}$) on each of our \textsc{comps} subsets. Chance performance is 50\% throughout.}
\label{tab:overall}
\vspace{-1.2em}
\end{table}

{\Cref{tab:overall} shows the average accuracy of the \plms{} on each subset of \comps{}. 
Despite the increase in complexity, we see that \plms{} {actually show slightly stronger} performance on \compswugs{} (68.9\%) than on \compsbase{} (67.1\%). This means that there are instances in which models prefer the property {in the positive context over the negative context (\ref{ex:acc3} > \ref{ex:unacc3}),}
but show the opposite behavior in \compsbase{} (\ref{ex:unacc4} > \ref{ex:acc4}). }
\ex.\label{ex:exwugs}
\small
    \a.\label{ex:acc3} A wug is a \textbf{robin}. Therefore, a wug can fly.
    \b.\label{ex:unacc3} A wug is a \textbf{penguin}. Therefore, a wug can fly.
    \c.\label{ex:acc4} A \textbf{robin} can fly.
    \d.\label{ex:unacc4} A \textbf{penguin} can fly.
    
{This pattern of performance could lead to spurious conclusions that models are successfully executing property inheritance, when in fact they show a lack of the pre-requisite property knowledge based on their failure on \compsbase{}}. {We will discuss these inconsistencies in more detail below. Overall, however, the relatively strong performance on \compswugs{} suggests} that models are largely unaffected when we control for {simple memorization of tested phrases---e.g., \textit{robin can fly}---by linking known concepts to properties through an intervening subordinate concept (\textit{wug})}. {This suggests that models are not relying on simple memorization, but does not control for the possibility of simple association between content words (\textit{robin} and \textit{fly})---for this we turn to \compswugsdist{}.}

{The \compswugsdist{} test assesses whether models retain strong property attribution performance when content words in the context are not all relevant for the property prediction.
The stimuli thus include irrelevant distractor concepts and their subordinates---which, in a robust model, should not affect attribution of the property to the correct concept} (see \cref{sec:stimuli} for stimulus construction).

From \Cref{tab:overall}, the average accuracies of \plms{} on both subsets of \compswugsdist{} (\befor{} and \inbetween{}) indicate that overall, models now {show clear degradation in property inheritance performance as a result of the distracting information}.
Specifically, the \plms{}' performance drops by 9.7 points on instances when the distracting information is added \befor{} the relevant context and queried property, and by 21.7 points on instances where it is added \inbetween{} the two, relative to the undistracted property inheritance stimuli (\compswugs{}).
Notably, the latter drop in performance brings models level with chance accuracy (we fail to reject the null hypothesis that avg.~accuracy of models is 50\%; $p = \text{.62}$, Wilcoxon signed rank exact text), highlighting a pronounced lack of robustness in \plms{}' capacity to attribute properties to the correct concepts in their input context.

\begin{figure}[!t]
    \centering
    \includegraphics[width=\columnwidth]{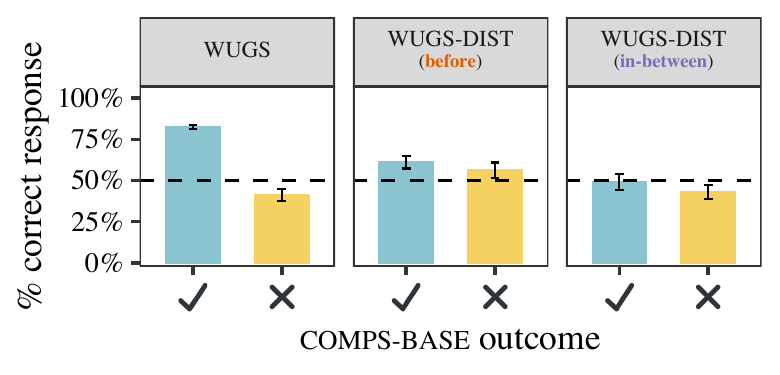}
    \caption{Distribution of model performance on \compswugs{} and \compswugsdist{} (both subsets) across possible outcomes (correct = \cmark{}, incorrect = \xmark{}) of the models on corresponding minimal pairs in \compsbase{}. Error bars indicate 95\% CI, while dashed line indicates chance performance (50\%).}
    \label{fig:wugsdeepdive}
    \vspace{-1.4em}
\end{figure}

\begin{figure*}[t!]
    \centering
    \includegraphics[width=\textwidth]{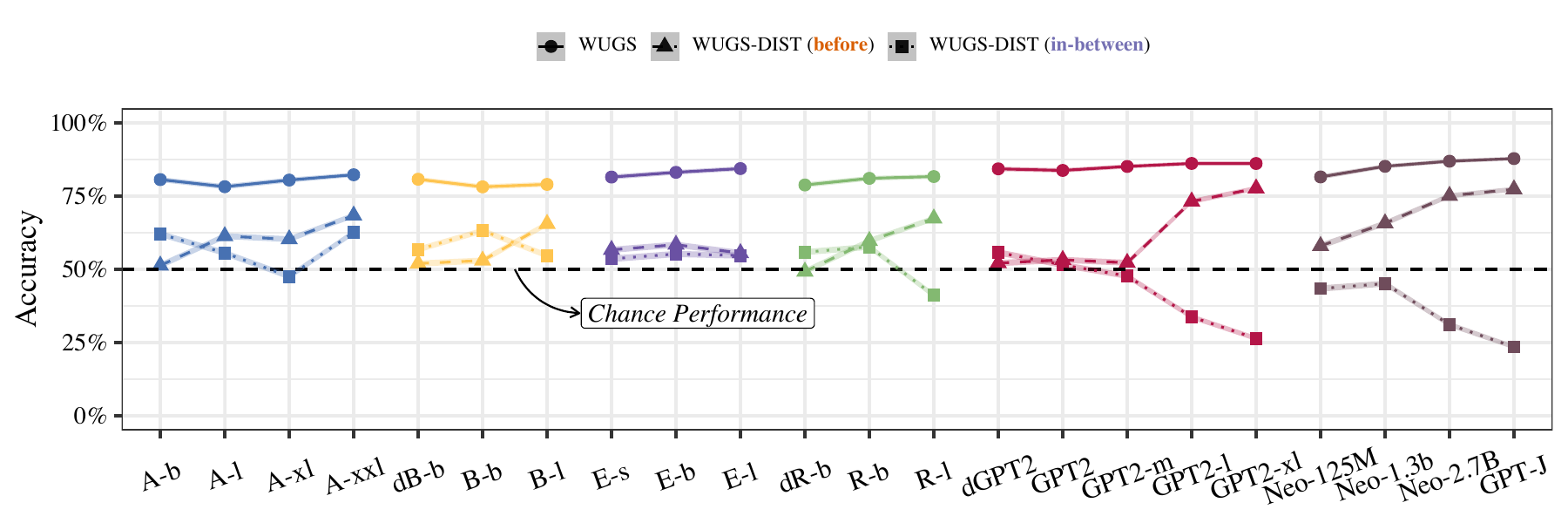}
    \caption{Accuracies of individual models (grouped by family, in increasing order based on number of parameters) on \compswugs{} and \compswugsdist{}. Black dashed line indicates chance performance (50\%). Refer to \Cref{tab:modelmeta} for unabbreviated model names. Error bands indicate 95\% Bootstrap CIs.}
    \label{fig:distraction}
    \vspace{-1em}
\end{figure*}

\paragraph{Accounting for spurious performance} 
The \compswugs{} results above {raise the concern that models are often showing} spurious performance: {accurately demonstrating property inheritance behavior without actually possessing the right property knowledge}. To shed more light on this potential issue, we plot the distribution of model accuracies on our property inheritance stimuli (\compswugs{} and \compswugsdist{}) {divided based on their outcomes on the corresponding} stimuli in \compsbase{}.
\Cref{fig:wugsdeepdive} shows these distributions. 
In \compswugs{} and both subsets of \compswugsdist{}, models show this spurious {correct} behavior on 41.3\%, 55.6\%, and 42.8\% of instances in which they produce incorrect judgments on the corresponding \compsbase{} stimuli (\plotyellow{} bars in \Cref{fig:wugsdeepdive}).
{This non-trivial proportion of cases with spurious performance further reinforces the idea that \plms{}' successful predictions on these tests are likely relying on heuristics rather than robust inferences about property knowledge.
We can remove the effects of these spurious instances by filtering to items in which models give the correct answer on \compsbase{} (\plotblue{} bars in \Cref{fig:wugsdeepdive})---though we see that the overall conclusions remain the same after this filtering.}

\paragraph{On the pronounced effect of proximity in autoregressive \plms{}} 
Our previous discussion summarized the aggregate property inheritance behavior of the 22 \plms{} we considered---we now zoom in for a model-wise analysis. \Cref{fig:distraction} shows models' relative accuracies on \compswugs{} and \compswugsdist{}, {filtering to items with correct \compsbase{} performance, as in the blue bars of \Cref{fig:wugsdeepdive}}.
Consistent with our overall findings, we observe distracting content to substantially degrade model performance across the board.\footnote{See also \citet{pandia-ettinger-2021-sorting} for a similar degradation of performance on cloze-tasks involving factual retrieval.}
A particularly noteworthy pattern is that the degradation in autoregressive PLM families---GPT2 and EleutherAI---shows a stark sensitivity to \textit{proximity effects}.
While these classes of model seem to suffer less when distracting content is added \befor{} the context containing the positive concept (thus placing the distraction farther from the queried property), they show substantially worse performance when the opposite is the case (i.e., when distraction is added \inbetween{}, and is therefore closer to the queried property). This degradation due to proximity of the distracting content becomes \textbf{catastrophically worse as models grow larger in the number of pre-trained parameters}---in fact bringing their performance down to as much as 26.2 points \textbf{below chance} (in GPT-J, which has 6B parameters).
While MLMs also show similar levels of degraded performance in presence of distraction, they do not seem to show any systematic sensitivity to proximity effects, likely due to their bidirectional nature.

\paragraph{Results on GPT-3}
In addition to our main experiments, we also evaluate GPT-3 \citep{brown2020language} models on a small subset of \comps{} stimuli (denoted as \minicomps{}). 
Results from this analysis (shown in \Cref{sec:gpt3}) are largely aligned with our main conclusions, with all GPT-3 models---including the largest one (175B parameters)---performing worse than chance on the \inbetween{} subset of \minicompswugsdist{}, while performing substantially better on \minicompswugs{} and \minicompswugsdist{} (\befor{}). 
Together with our main results, this indicates that scaling alone may be insufficient to elicit robust inferences about concepts and their properties.

\paragraph{Choice of nonce words} 
Nonce words constitute an important design decision for our stimuli---we followed precedents in language acquisition research \citep[\textit{i.a.}]{berko1958child, gopnik2000detecting} and used previously existing nonce words (such as \textit{wug} and \textit{blicket}) to represent novel concepts in context. While these are expected to be novel for humans, they may appear in pre-training corpora on which \plms{} are usually trained.\footnote{e.g., \textit{wug} appears in wikipedia: \url{https://en.wikipedia.org/wiki/Jean_Berko_Gleason} (accessed on Jan 23)} 
This raises a potential concern that \plms{} could already be biased toward certain properties for these words (e.g., \textit{wug} is commonly depicted as a bird), and may struggle to associate them with different properties.\footnote{We thank Reviewers 1 and 3, Najoung Kim and Kyle Mahowald for raising this concern.}
To explore this empirically, we conducted experiments with alternative nonce words (generated synthetically, similar to \citet{kim2022uncontrolled}; see  \Cref{sec:nonce}).
\Cref{fig:distractionauto} (\Cref{sec:nonce}) shows results on \compswugs{} and \compswugsdist{} with randomly sampled nonce words. {We see that the new results are comparable to those in \Cref{fig:distraction}, with models showing the same preference on both stimulus versions 80\% of the time on average. This suggests that the choice of nonce words is not producing any noteworthy bias.}

\paragraph{Framing of novel taxonomic information} Another relevant stimulus design decision is the phrasing for introducing novel concepts in context. While we used \textit{``A wug is a \texttt{[CONCEPT]}''} for our main experiments, we additionally tested with an alternate framing: \textit{``A wug is a type of \texttt{[CONCEPT]}.''}
From \Cref{fig:distractiontypeof} (\Cref{sec:alternate}), we again see that the overall patterns of results {are comparable} to the original results, with models showing the same preference {across both versions of the stimuli} on \compswugs{} and \compswugsdist{} 90\% of the time, on average.

\section{General Discussion and Conclusion}
The overall goal of \comps{} is to shed light on the extent to which \plms{} can robustly (1) attribute to real world concepts (e.g., \textsc{horse}, \textsc{whale}) their {correct} properties (e.g., \textit{is a mammal}); and (2) demonstrate behavior consistent with \textit{property inheritance}: a reasoning process in which concepts are endowed with the properties of their superordinates \citep{smith1978theories, sloman1998categorical, murphy2004big}. 
Testing \plms{} for these abilities allows us to ask key questions about
how they encode and transfer knowledge.
To target these capabilities more precisely, and mitigate potential inflation of performance by superficial heuristics such as coarse-grained similarity and word association, we propose incrementally increasing levels of controls in constructing our minimal pair stimuli, progressively making the task of attributing properties to concepts more challenging.

Findings from our initial experiment on \compsbase{} established that the basic capacity of models to attribute properties to everyday concepts is largely coarse grained. \plms{} were more successful in making correct property attributions when the candidate concepts were radically different, and struggled when the concepts shared  semantic relations or had high co-occurrence. On testing for `property inheritance' behavior (via \compswugs{}), \plms{} initially appeared to demonstrate reasonable success, but they also showed spurious behavior in achieving correct performance on a non-trivial number of instances for which they did not succeed in the prerequisite base condition. 
Furthermore, this performance declined substantially in {the presence} of distracting information (\compswugsdist{}), {providing further evidence that what property knowledge and reasoning we appear to see in these \plms{} is more reliant on superficial heuristics than on ideal reasoning behavior}. Of particular note is our finding of \emph{catastrophic distraction} in large autoregressive \plms{}, whose sensitivity to proximity effects brings their overall performance well under chance, especially when scaled up to billions of parameters.

Contemporary work has highlighted the promise of \plms{} on high-level tasks requiring---among other things---access to proper relational knowledge between concepts \citep[see][]{petroni2019language, safavi-koutra-2021-relational, piantadosi2022meaning}.
By drawing on the concept of property inheritance, our experiments target reasoning ability based on perhaps the most well-established of relations---the taxonomic or the \texttt{IsA} relation \citep{murphy2003semantic}.
Recent work has also alluded to the proficiency of \plms{} in capturing taxonomic information about everyday objects and entities \citep[though see \citet{ravichander-etal-2020-systematicity}]{weir2020probing, chen-etal-2021-constructing}.
Findings from our controlled experiments suggest that 
\plms{}' approximation of the consequences of the taxonomic relation is at best noisy, {in light of} clear failures especially in presence of similarity-governed competition.
We conclude from our analyses that instead of robustly extracting relational information and reasoning about properties of concepts, it is likely that the \plms{} tested here are optimized to prefer superficial cues in making word predictions, leading to mistakes and inaccuracies in presence of irrelevant and distracting information. 
Since robust natural language understanding will be critically reliant on understanding of property knowledge and implications of property transfer, we hope that these findings will motivate adoption of rigorous assessment methods as well as work toward more robust property knowledge and reasoning in \plms{}.

\section*{Limitations}
\paragraph{Zero-shot setup} 
Using a zero-shot setup to test \plms{} for human-like capacities such as property inheritance (as we have done in this work) has recently come under scrutiny.
In particular, \citet{lampinen2022recursive} argues that such a setup could be problematic because \plms{} are trained to imitate the language produced by countless individuals with different beliefs, cultures, and behaviors. As a result, \plms{} are likely to be handicapped in assigning sufficient probability mass to the desired family of continuations, given minimal prompts without any particular task-specific context.
Instead, \citet{lampinen2022recursive} suggests the need for \plms{} 
\begin{quote}
    ``[...] to be guided into an experiment-appropriate behavioral context, analogously to the way cognitive researchers place humans in an experimental context, and orient them toward the task with instructions and examples.''
\end{quote} 
This criticism is valid, and it is possible that models could overcome their lack of robustness to distraction effects by observing examples of our stimuli in context, though this has largely been shown in \plms{} that are significantly larger than the ones we have tested in this work \citep{brown2020language, chowdhery2022palm, wei2022emergent}.\footnote{though see recent work by \citet{shi2023large}, who show distraction effects in such large \plms{} in solving arithmetic reasoning problems, even after using sophisticated in-context prompting methods such as Chain-of-Thought \citep{wei2022chain}, Least-to-Most \citep{zhou2022least}, and Self-Consistency \citep{wang2022self}.}
Indeed, recent work has demonstrated these larger \plms{} to achieve strong performance on other types of reasoning---such as those required for solving math problems, reversing sequences, etc.---by priming models to produce additional textual content that represents intermediate reasoning steps and explanations \citep{nye2021show, wei2022chain, lampinen2022can}, in a few-shot setting.\footnote{See also \citet{sinha2022language}, who analyze \plms{} comparable in size to those studied in this work in a few-shot minimal-pair setting.}
At the same time, a few-shot version of \comps{} stimuli could expose models to the possibility of leveraging heuristics that are naturally absent in the zero-shot setup, and therefore such a setup would critically require the design of additional controls, which we leave for future work.
\paragraph{Ideal reasoning behavior} Another limitation of our work is that it takes ideal and robust property inheritance behavior as the monolithic gold-standard for human cognition, something that recent work has cautioned against \citep{pavlick2019inherent, dasgupta2022language, webson2023language}.
Although we relied on a database of concept-property pairs that were largely generated by human participants, whether or not humans will be robust to the types of distraction that were observed in \plms{} is an open question and requires further investigation.
However, notably we are not making direct comparisons between models and humans here---we argue that our primary contribution of controlled stimuli that tease apart shallow processing from robust conceptual reasoning in \plms{} bears substantial merit that is independent from any comparisons between humans and computational systems. 
{Furthermore, we emphasize that we are setting a reasonable---and to a certain extent, human-independent---desideratum in this work, which is that models should robustly capture ground-truth knowledge about everyday concepts and their properties and reflect this knowledge in their inferences about newly introduced concepts.}

\paragraph{Behavioral evaluation} This work tests and analyses \plms{} on property knowledge and property inheritance only from a behavioral perspective, which at its core is a correlational endeavor. Potential future work could complement our results by providing evidence from representational analyses, or by devising causal interventions, similar to those recently explored in the realm of syntactic agreement \citep{finlayson-etal-2021-causal}, or in testing of negation and hypernymy in NLI models \citep{geiger-etal-2020-neural}, among others. Importantly, this would require the development of new methods that shed light on how new information---such as the ones we use in \compswugs{} and \compswugsdist{}---is integrated into the model (see \citet{misra2022property} for an example of such an analysis for novel properties).

\paragraph{Targeted language} Finally, \comps{} only consists of sentences in English, thereby biasing our results only for \plms{} trained in that language. 

\section*{Acknowledgments}
For helpful comments we thank Najoung Kim, Tal Linzen, Brenden Lake, Kyle Mahowald, Andrew Lampinen, the anonymous reviewers, and audiences at the Computation and Psycholinguistics Lab (NYU), the Human and Machine Learning Lab (NYU), the UChicago CompLing Lab, UT Austin Linguistics Grad Student Seminar
, and the MIT Department of Brain and Cognitive Sciences. Any errors are our own. Our experiments were conducted with resources provided by the Rosen Center for Advanced Computing at Purdue University \citep{McCartney2014}. We are also grateful to Sam Huang for helping out with experiments conducted on GPT-3/3.5 models.

\bibliography{custom, additional}
\bibliographystyle{acl_natbib}

\appendix


\section{Model Metadata}
\label{sec:modelmeta}
\Cref{tab:modelmeta} shows the different models used in our experiments, along with their abbreviation, tokenization scheme, total parameters, vocabulary size, number of tokens encountered during training, and corpora on which they are pre-trained.
All models were accessed using 
\texttt{minicons} \citep{misra2022minicons},\footnote{\url{https://github.com/kanishkamisra/minicons}} a python library that serves as a wrapper around
Huggingface's \texttt{transformers} \citep{wolf-etal-2020-transformers}, 
and provides a unified mechanism for eliciting log-probabilities in batch-wise manner for any autoregressive or masked LM that is accessible through the huggingface hub, or is trained using the transformers library.
Experiments were performed using an NVIDIA V100 GPU (32 GB RAM) and took about 6 hours to run, discounting the time it took to download the models from the Huggingface Hub.\footnote{\url{https://huggingface.co/models}.}




\begin{table*}[!ht]
\centering
\resizebox{\linewidth}{!}{%
\begin{tabular}{clrrclr} 
\toprule
\textbf{Family} & \textbf{Model (Abbrev.)} & \textbf{Parameters} & \textbf{Vocab Size} & \textbf{Tokenization} & \textbf{Corpora} & \textbf{Tokens} \\ 
\midrule
\multirow{4}{*}{\textcolor[HTML]{2e59a8}{ALBERT}} & \texttt{albert-base-v2} (A-b) & 11M & \multirow{4}{*}{30,000} & \multirow{4}{*}{SentencePiece} & \multirow{4}{*}{\textsc{Wiki} and \textsc{bc}} & \multirow{4}{*}{3.3B} \\
 & \texttt{albert-large-v2} (A-l) & 17M &  &  &  &  \\
 & \texttt{albert-xlarge-v2} (A-xl) & 59M &  &  &  &  \\
 & \texttt{albert-xxlarge-v2} (A-xxl) & 206M &  &  &  &  \\ 
\midrule
\multirow{3}{*}{\textcolor[HTML]{fe9929}{BERT}} & \texttt{distilbertbase-uncased} (dB-b) & 67M & \multirow{3}{*}{30,522} & \multirow{3}{*}{WordPiece} & \multirow{3}{*}{\textsc{Wiki} and \textsc{bc}} & \multirow{3}{*}{3.3B} \\
 & \texttt{bert-base-uncased} (B-b) & 110M &  &  &  &  \\
 & \texttt{bert-large-uncased} (B-l) & 345M &  &  &  &  \\ 
\midrule
\multirow{3}{*}{\textcolor[HTML]{54278f}{ELECTRA}} & \texttt{electra-small} (E-s) & 13M & \multirow{3}{*}{30,522} & \multirow{3}{*}{WordPiece} & \multirow{2}{*}{\textsc{Wiki} and \textsc{bc}} & \multirow{2}{*}{3.3B} \\
 & \texttt{electra-base} (E-b) & 34M &  &  &  &  \\
 & \texttt{electra-large} (E-l) & 51M &  &  & \begin{tabular}[c]{@{}l@{}}\textsc{Wiki}, \textsc{bc}, \textsc{cw},\\\textsc{cc}, and \textsc{Giga}\end{tabular} & 33B \\ 
\midrule
\multirow{3}{*}{\textcolor[HTML]{238443}{RoBERTa}} & \texttt{distilroberta-base} (dR-b) & 82M & \multirow{3}{*}{50,265} & \multicolumn{1}{l}{\multirow{3}{*}{Byte-pair encoding}} & \textsc{owtc} & 2B \\
 & \texttt{roberta-base} (R-b) & 124M &  &  & \multirow{2}{*}{\begin{tabular}[c]{@{}l@{}}\textsc{bc}, \textsc{cc-news},\\\textsc{owtc}, and \textsc{Stories}\end{tabular}} & \multirow{2}{*}{30B} \\
 & \texttt{roberta-large} (R-l) & 355M &  & \multicolumn{1}{l}{} &  &  \\ 
\midrule
\multirow{6}{*}{\textcolor[HTML]{93003a}{GPT2}} & \texttt{distilgpt2} (dGPT2) & 82M & 50,257 & \multirow{5}{*}{Byte-pair encoding} & \textsc{owtc} & 2B \\
 & \texttt{gpt2} (GPT2) & 124M & \multirow{4}{*}{50,257} &  & \multirow{4}{*}{\textsc{WebText}} & \multirow{4}{*}{8B$^*$} \\
 & \texttt{gpt2-medium} (GPT2-m) & 355M &  &  &  &  \\
 & \texttt{gpt2-large} (GPT2-l) & 774M &  &  &  &  \\
 & \texttt{gpt2-xl} (GPT2-xl) & 1.5B &  &  &  &  \\ 
\midrule
\multirow{4}{*}{\textcolor[HTML]{6f4c5b}{EleutherAI}} & \texttt{gpt-neo-125M} (Neo-125M) & 125M & \multirow{4}{*}{50,257} & \multirow{4}{*}{Byte-pair encoding} & \multirow{4}{*}{\textsc{pile}} & 300B \\
 & \texttt{gpt-neo-1.3B} (Neo-1.3B) & 1.3B &  &  &  & 380B \\
 & \texttt{gpt-neo-2.7B} (Neo-2.7B) & 2.7B &  &  &  & 420B \\
 & \texttt{gpt-j-6B} (GPT-J) & 6B &  &  &  & 402B \\
\bottomrule
\end{tabular}
}
\caption{Summary of the 22 models that we evaluate in this paper. \textbf{Legend for Corpora:} \textsc{Wiki}: Wikipedia; \textsc{bc}: BookCorpus \citep{zhu2015aligning}; \textsc{cw}: ClueWeb \citep{callan2009clueweb09}; \textsc{cc}: CommonCrawl \textsc{Giga}: Gigaword \citep{graff2003english}; \textsc{owtc}: OpenWebTextCorpus \citep{Gokaslan2019OpenWeb}; \textsc{cc-news}: CommonCrawl News \citep{nagel2016ccnews}; \textsc{Stories}: Stories corpus \citep{trinh2018simple}; \textsc{WebText}: WebText corpus \citep{radford2019language}; \textsc{Pile}: The Pile \citep{gao2020pile}.\\$^*$As estimated by \citet{warstadt-etal-2020-blimp-benchmark}.}
\label{tab:modelmeta}
\end{table*}

\section{Preview of \textsc{comps} stimuli}
\label{sec:appstimuli}

\def\arraystretch{1.15}
\begin{table*}[t!]
\centering
\begin{tabular}{@{}llr@{}}
\toprule
\textbf{\comps{} subset} & \textbf{Stimulus} & \textbf{Score} \\ \midrule
\multirow{2}{*}{\textsc{base}} & A \textbf{horse} has hooves. & -3.829 \\
 & A \textbf{dog} has hooves. & -4.963 \\ \midrule
\multirow{2}{*}{\textsc{wugs}} & A fep is a \textbf{horse}. Therefore, a fep has hooves. & -2.153 \\
 & A fep is a \textbf{dog}. Therefore, a fep has hooves. & -3.392 \\ \midrule
\multirow{2}{*}{\textsc{wugs-dist} (\befor{})} & A wug is a dog. A fep is a horse. Therefore, a \textbf{fep} has hooves. & -2.919 \\
 & A wug is a dog. A fep is a horse. Therefore, a \textbf{wug} has hooves. & -2.895 \\ \midrule
\multirow{2}{*}{\textsc{wugs-dist} (\inbetween{})} & A fep is a horse. A wug is a dog. Therefore, a \textbf{fep} has hooves. & -3.616 \\
 & A fep is a horse. A wug is a dog. Therefore, a \textbf{wug} has hooves. & -3.092 \\ \bottomrule
\end{tabular}%
\caption{An example of matched stimuli across different \comps{} subsets, as well as conditional log-probabilities elicited by GPT-J. Here, the property of interest is \textit{has hooves}, the positive concept is \textsc{horse}, and the negative concept is \textsc{dog}. The negative concept in this case was sampled using the co-occurrence knowledge representation method (see \Cref{sec:compsneg}). Emboldened words indicate items that are different in the minimal pair. Refer to \Cref{sec:performance} for discussion on how `Score' is computed.}
\label{tab:examplescores}
\end{table*}
We show examples of stimuli from our \compsbase{}, \compswugs{}, and \compswugsdist{} datasets in \Cref{lst:compsbase} and \Cref{lst:compswugs}, respectively.
Stimuli with distraction---i.e., in \compswugsdist{}---are similar to that in \Cref{lst:compsbase}, but with the \texttt{distraction\_type} value set to either \texttt{`before'} or \texttt{`in-between'}.

\begin{listing}[!ht]
\begin{minted}{js}
{
    "id": 12706, 
    "property": "can fly", 
    "acceptable_concept": "owl", 
    "unacceptable_concept": "squirrel", 
    "prefix_acceptable": "an owl", 
    "property_phrase": "can fly.", 
    "prefix_unacceptable": "a squirrel", 
    "condition": "co-occurrence", 
    "similarity": 0.62
}
\end{minted}
\caption{An instance of \compsbase{}. \texttt{``condition''} represents the negative sampling scheme, and \texttt{``similarity''} represents the similarity between the acceptable concept and the unacceptable concept on the basis of the condition (either Taxonomic, Property Norm, Co-occurence, or Random).}
\label{lst:compsbase}
\end{listing}

\begin{listing}[!ht]
\begin{minted}[breaklines]{js}
{
    "item": 8343, 
    "comps_id": 28798, 
    "property": "has hooves", 
    "acceptable_concept": "horse", 
    "unacceptable_concept": "dog", 
    "prefix_acceptable": "A dax is a horse. Therefore, a dax", 
    "prefix_unacceptable": "A dax is a dog. Therefore, a dax", 
    "property_phrase": "has hooves.", 
    "negative_sample_type": "co-occurrence", 
    "similarity": 0.62, 
    "distraction_type": "undistracted"
}
\end{minted}
\caption{An instance of \compswugs{}. \texttt{``condition''}  and \texttt{``similarity''} are the same as in \Cref{lst:compsbase}. \texttt{``distraction\_type''} denotes the type of distraction used (undistracted, before, in-between).}
\label{lst:compswugs}
\end{listing}

\noindent
\Cref{tab:examplescores} shows examples from each subset of \comps{}, and the conditional log-probability scores as computed by GPT-J \citep{gpt-j}, the largest LM tested on the full set of stimuli.

\section{Additional findings and analyses}
\label{sec:extra}

\subsection{Testing GPT-3/3.5}
\label{sec:gpt3}
Recent work in scaling \plms{} to hundred billion parameters has led to models such as GPT-3 \citep{brown2020language}, which are significantly larger than the largest model tested in the results discussed above (i.e., GPT-J, with 6B parameters).
Testing them on the entire set of \comps{} stimuli (49K $+$ 3 $\times$ 13.8K pairs of sentences) is prohibitively expensive since they are only accessible through paid APIs. 
Nonetheless, we sampled a small set of \comps{} stimuli---which we term as \minicomps{}---in order to get a glimpse of how well substantially larger \plms{} elicit property knowledge and demonstrate reasoning behavior compatible with property inheritance.
Specifically, we created \minicomps{} by sampling 1200 minimal pairs from each of our original \comps{} subsets (matched in terms of real world concepts and properties across the subsets), such that all pairs of nonce words in the resulting \minicompswugsdist{} end up being sampled equal number of times (100 times each).

\paragraph{Models} As test subjects, we chose four GPT-3 models \citep{brown2020language}: \texttt{ada}, \texttt{babbage}, \texttt{curie}, \texttt{davinci}, with the last one being the largest (at 175B parameters), and an additional fifth \texttt{davinci}-based model called \texttt{text-davinci-001}, which fine-tunes \texttt{davinci} on human-written demonstrations. We also test the recently proposed GPT-3.5 models, \texttt{text-davinci-002} and \texttt{text-davinci-003}, which improve over \texttt{davinci} by additionally fine-tuning it on code \textit{and} human-written demonstrations \citep{ouyang2022training}.\footnote{These models are also known as InstructGPT, as discussed in \url{https://platform.openai.com/docs/model-index-for-researchers}.} 
All these models are autoregressive in nature, so we use the same scoring and evaluation method as described in \Cref{sec:performance}.
Since the four original GPT-3 models (\texttt{ada}, \texttt{babbage}, \texttt{curie}, \texttt{davinci}) are trained using the same LM objective on the same corpora, we analyze them separately from \texttt{text-davinci-001}, \texttt{text-davinci-002}, and \texttt{text-davinci-003}, which we only compare to \texttt{davinci}. 
We do this to remain consistent with the way we displayed results in \Cref{sec:experiments}---ordering models based on their number of trained parameters---and also because models in the \texttt{text-davinci-XXX} series use the same underlying \texttt{davinci} model augmented with additional training mechanisms (e.g., reinforcement learning and fine-tuning on human-feedback) and data (e.g., code) instead of increasing its size, to our knowledge.

\begin{figure}[t!]
    \centering
    \includegraphics[width=\columnwidth]{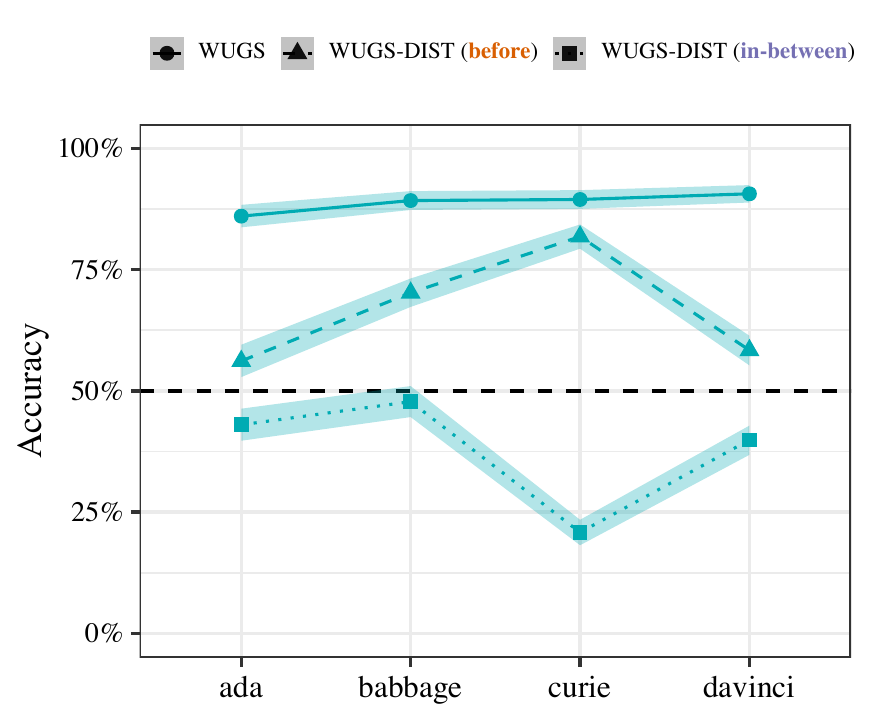}
    \caption{Accuracies of GPT-3 models (arranged in increasing order of the number of trained parameters) on \minicompswugs{} and \minicompswugsdist{}. Black dashed line indicates chance performance (50\%). Error bands indicate 95\% Bootstrap CIs. }
    \label{fig:gpt3distraction}
\end{figure}

\begin{figure}[t!]
    \centering
    \includegraphics[width=\columnwidth]{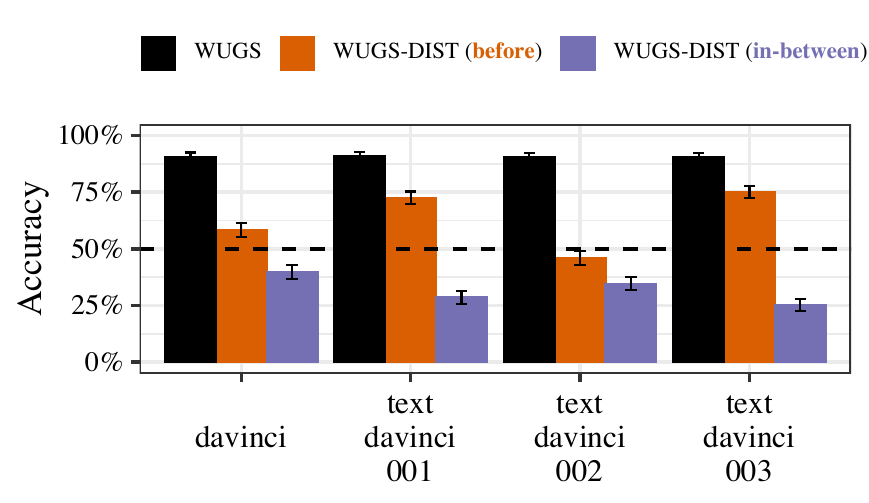}
    \caption{Accuracies of \texttt{davinci} models (GPT-3 and GPT-3.5)  on \minicompswugs{} and \minicompswugsdist{}. Black dashed line indicates chance performance (50\%). Error bars indicate 95\% Bootstrap CIs. \texttt{davinci} and \texttt{text-davinci-001} are GPT-3 \citep{brown2020language} models, while \texttt{text-davinci-002} and \texttt{text-davinci-003} are GPT-3.5 models.}
    \label{fig:davincidistraction}
\end{figure}

\paragraph{Results} \Cref{fig:gpt3distraction} shows the performance of the four GPT-3 models on \minicompswugs{} and \minicompswugsdist{}, while \Cref{fig:davincidistraction} compares GPT-3 \texttt{davinci} to its code and human-feedback adapted counterparts.
From \Cref{fig:gpt3distraction}, we see robustness issues to persist even for GPT-3 models, similar to our main results. 
Models perform remarkably well in the absence of distraction (i.e., on \minicompswugs{}), but struggle in its presence, especially when it is closer to the queried property. 
In particular, performance on \minicompswugsdist{} (\befor{}) increases with an increase in parameters until the largest model (\texttt{davinci}), where the performance drops closer to chance. On \minicompswugsdist{} (\inbetween{}), all models perform catastrophically worse than chance. 
This noteworthy pattern of proximity-based degradation in performance mimics the results shown in \Cref{fig:distraction}, though we do not see a systematic decline in performance with an increase in parameters as observed in the GPT2 and EleutherAI models---with the 175B parameter model (\texttt{davinci}) demonstrating an increase in performance over the relatively smaller \texttt{curie} model.

While the above results demonstrate that simply scaling autoregressive \plms{} is unlikely to overcome the lack of robustness against distracting content, we now test whether augmenting these large \plms{} by additionally training on code (GPT-3.5 models) and aligning them with human-provided demonstrations (\texttt{text-davinci-001} and both GPT-3.5 models) could lead to any improvements.
For instance, training on code could provide training signals to \plms{} that encourage entity tracking, which could potentially enable them, in our case, to resolve which subordinate concept (e.g., \textit{wug} vs. \textit{dax}) the target property is more likely to be associated with. 
Similarly, aligning with human-written demonstrations could potentially improve their truthfulness, which in our case, could lead to them to prefer correct property assignments.
However, from \Cref{fig:davincidistraction}, we see no noteworthy improvements demonstrated by these augmented models. 
All augmented models achieved similar accuracies on \compswugs{} as the \texttt{davinci} model (within 90.5\% and 91\%), suggesting that their augmentations preserved the general associations between the lexical items that denote everyday concepts and properties.
On stimuli containing distraction (i.e., both subsets of \compswugsdist{}), either the models performed systematically worse as compared to \texttt{davinci} (with \texttt{text-davinci-002} showing below-chance performance on both subsets), or they showed mixed results, where an improvement on \compswugsdist{} (\befor{}) was accompanied by a decline on \compswugsdist{} (\inbetween{}).

Together, these results suggest that neither an increase in scale nor additional training methods such as alignment with human instructions/feedback or training on code prevents models from being distracted in associating properties to novel subordinate concepts introduced in the input context. In fact, the catastrophic effects of proximity-based distraction persists even for the most recent state of the art GPT-3/3.5 models.

\begin{figure*}[t!]
    \centering \includegraphics[width=\textwidth]{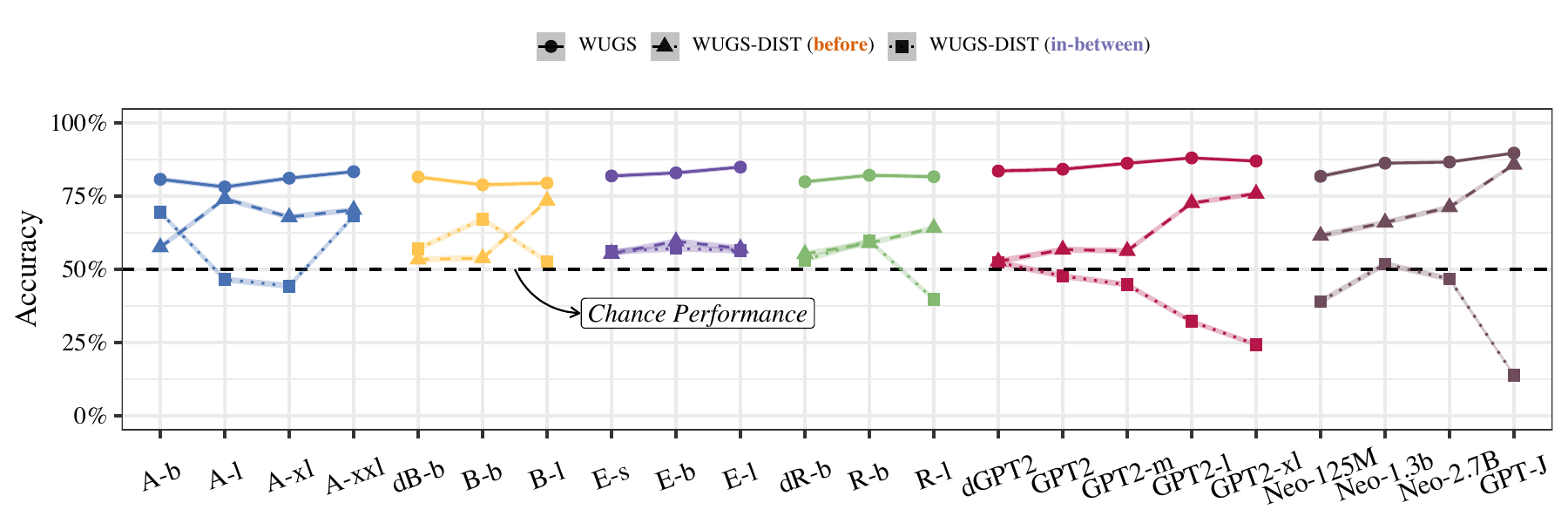}
    \caption{Accuracies of individual models (grouped by family, in increasing order based on number of parameters) on \compswugs{} and \compswugsdist{} with \textbf{synthetically constructed nonce words}. Black dashed line indicates chance performance (50\%). Refer to \Cref{tab:modelmeta} for unabbreviated model names.}
    \label{fig:distractionauto}
\end{figure*}

\begin{figure*}[ht!]
    \centering
    \includegraphics[width=\linewidth]{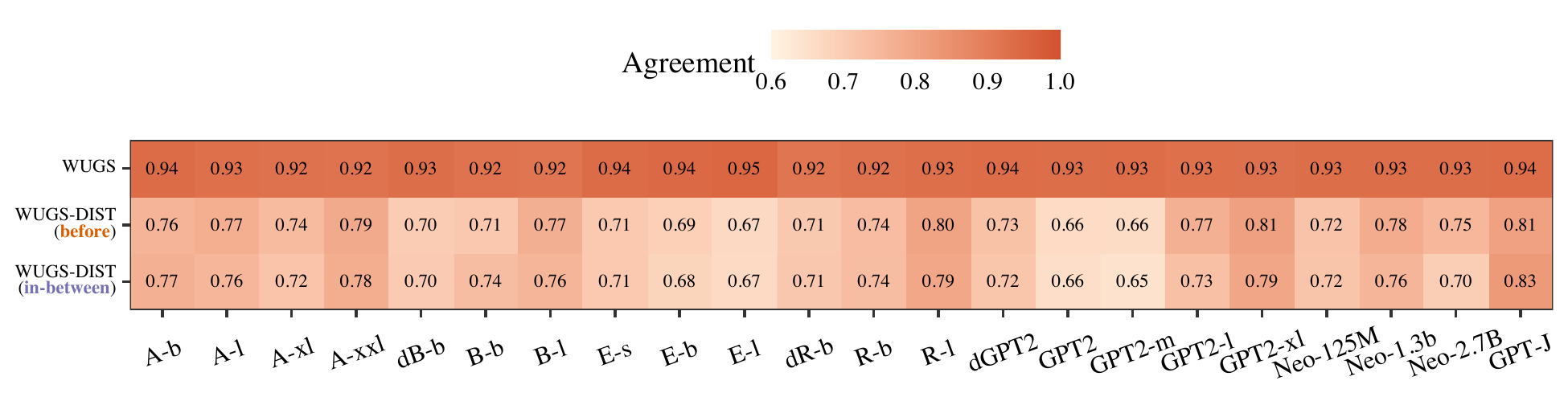}
    \caption{Proportion of cases (in \compswugs{} and both subsets of \compswugsdist{}) \textbf{where each listed model's preference on the original stimuli matches that in stimuli with synthetically constructed nonce words}, measured as the `Agreement'. An agreement of 1.0 suggests that a given model's preferences are perfectly matched across both sets of stimuli.}
    \label{fig:agreement}
\end{figure*}

\subsection{Results with alternate nonce words}
\label{sec:nonce}
Here we report results on \compswugs{} and \compswugsdist{} using an alternate set of nonce words, which we constructed by sampling (with replacement) from 26 lower-case ASCII alphabet characters. Specifically, we constructed novel character sequences---each assigned as a replacement for our original four nonce words---of lengths ranging from 4-8 by sampling in an alternate fashion from consonants (odd positions) and vowels (even positions).\footnote{the resulting set of words is: \{\textit{ruhisin, kifosa, rosibif, lepuvu}\}, still amounting to 12 unique ordered pairs in the \compswugsdist{} stimuli.}
A replication of \Cref{fig:distraction} using the stimuli with these newly sampled nonce words is shown in \Cref{fig:distractionauto}.
On comparing figures \ref{fig:distraction} and \ref{fig:distractionauto}, we observe largely similar patterns of results on stimuli containing nonce words constructed using randomly sampled characters. That is, models generally performed well on \compswugs{}, while they struggled on \compswugsdist{}. 
There were some exceptions: (1) GPT-Neo 1.3B and 2.7B showed improvements (relative to the original stimuli) in cases where distraction is added closer to the queried property (i.e., \inbetween{}), though they still hover around chance performance, and additionally the performance of GPT-J, like in the original results is still substantially below chance; and (2) there were non-trivial improvements demonstrated by ALBERT models (large and xl) on the \befor{} subset of \compswugsdist{}, and BERT-large on the \inbetween{} subset of \compswugsdist{}.

\begin{table}[t!]
\centering
\resizebox{\columnwidth}{!}{%
\begin{tabular}{@{}lc@{}}
\toprule
\textbf{Stimuli} & \textbf{Avg. Agreement} \\ \midrule
\compswugs{} & 93.1$_{\text{0.8}}$ \\
\compswugsdist{} (\befor) & 73.9$_{\text{4.6}}$ \\
\compswugsdist{} (\inbetween) & 73.0$_{\text{4.6}}$ \\ \midrule
Overall & 80.0$_{\text{2.9}}$ \\ \bottomrule
\end{tabular}%
}
\caption{Average agreement ($\times$ 100) in \plms{}' preference on stimuli containing original and synthetically constructed nonce words.}
\label{tab:agreement}
\vspace{-1em}
\end{table}

To precisely quantify the difference between the two sets of results, we measured the agreement between the predictions of the \plms{} for both sets of stimuli, taken as the proportion of minimal pairs in which the models' relative preference agree.

\Cref{fig:agreement} shows individual model agreement on \compswugs{} and \compswugsdist{}, while \Cref{tab:agreement} shows agreement percentages averaged across all models. From these results we observe models to show greater robustness to the variability introduced by the choice of nonce words in stimuli with one novel concept (\compswugs{}) than in stimuli with multiple novel concepts (\compswugsdist{}). 
Despite this discrepancy, there is generally a high average agreement (80\%) between a given model's set of decisions on stimuli with original and alternative nonce words.

\subsection{Results with alternate templates}
\label{sec:alternate}

Here we report results on an alternate phrasing of our stimuli, where instead of using the original template for introducing novel concepts in context (\textit{a wug is a \texttt{[CONCEPT]}}), we use: \textit{A wug is a type of \texttt{[CONCEPT]}}, where \textit{wug} indicates the novel concept. In all cases, we simply alter the template, keeping everything else constant, including the choice of nonce words.

\begin{figure*}[t]
    \centering
\includegraphics[width=\linewidth]{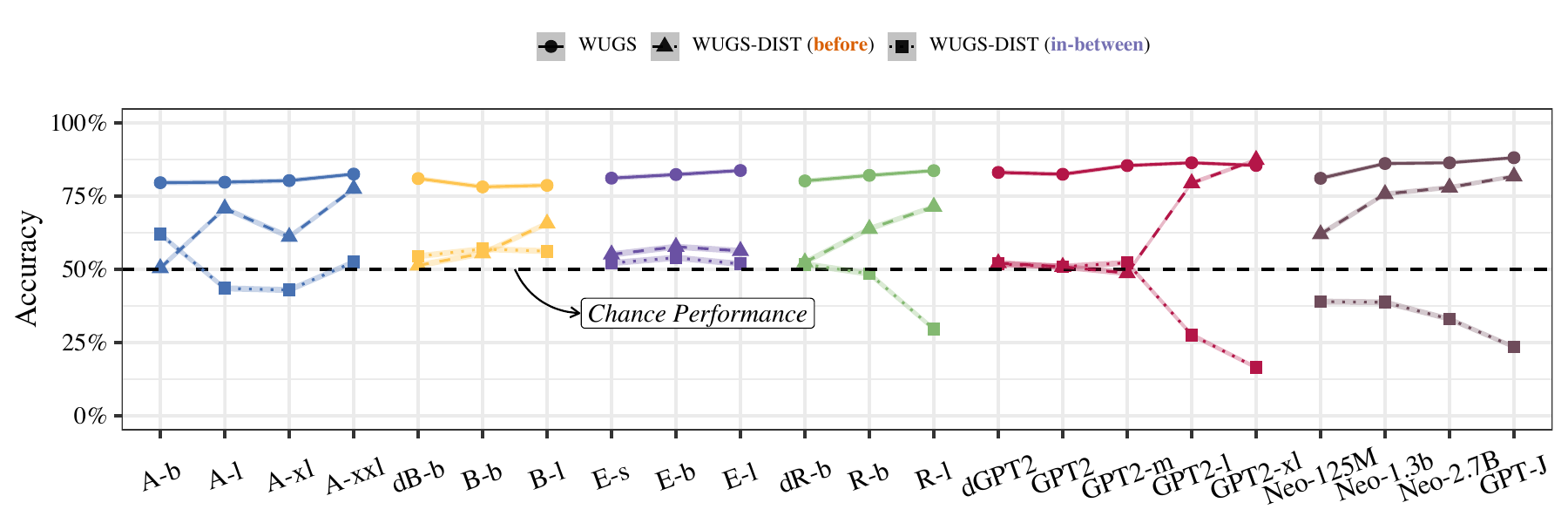}
    \caption{Accuracies of individual models (grouped by family, in increasing order based on number of parameters) on \compswugs{} and \compswugsdist{} with \textbf{alternate framing of novel taxonomic information}. Black dashed line indicates chance performance (50\%). Refer to \Cref{tab:modelmeta} for unabbreviated model names.}
    \label{fig:distractiontypeof}
\end{figure*}

\begin{figure*}[t]
    \centering
    \includegraphics[width=\linewidth]{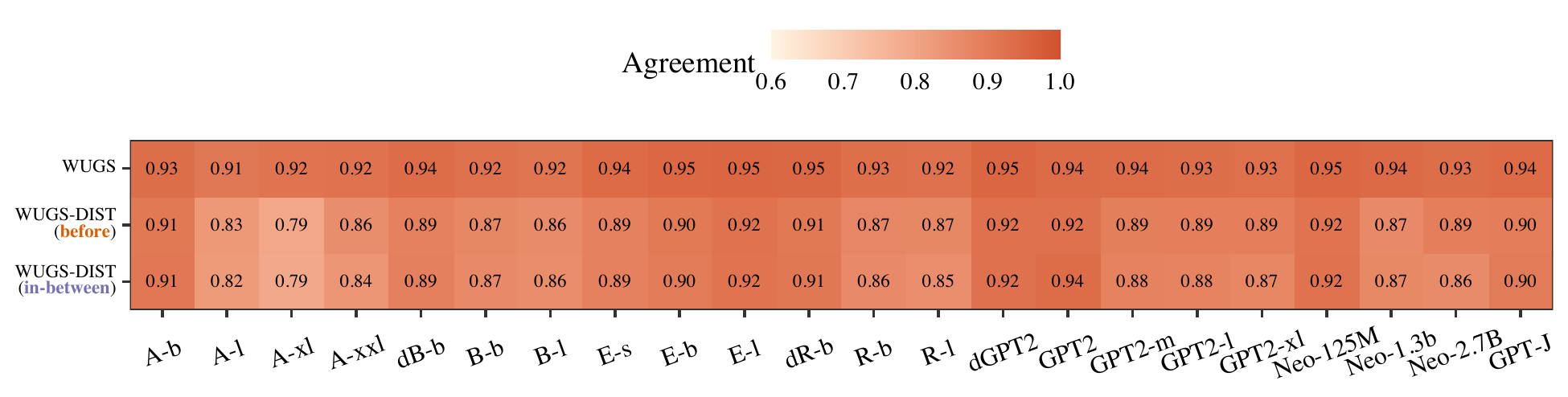}
    \caption{Proportion of cases (in \compswugs{} and both subsets of \compswugsdist{}) \textbf{where each listed model's preference on the original stimuli matches that in stimuli with alternate framing of novel taxonomic information}, measured as the `Agreement'. An agreement of 1.0 suggests that a given model's preferences are perfectly matched across both sets of stimuli.}
    \label{fig:individualagreementtypeof}
\end{figure*}

\Cref{fig:distractiontypeof} shows accuracies of the models on stimuli with this alternate phrasing, while \Cref{fig:individualagreementtypeof} and \Cref{tab:agreementtypeof} show individual and averaged overall agreement between models' preference on original and the alternatively-phrased stimuli, respectively. The agreement percentages between models' preferences are quite high (average agreement being 90\%)---in fact even greater than the agreement observed as a result of altering the nonce words (\Cref{tab:agreement}), further cementing the robustness of our results.

\begin{table}[t!]
\centering
\resizebox{\columnwidth}{!}{%
\begin{tabular}{@{}lc@{}}
\toprule
\textbf{Stimuli} & \textbf{Avg. Agreement} \\ \midrule
\compswugs{} & 93.4$_{\text{1.2}}$ \\
\compswugsdist{} (\befor) & 88.5$_{\text{3.1}}$ \\
\compswugsdist{} (\inbetween) & 88.0$_{\text{3.6}}$ \\ \midrule
Overall & 90.0$_{\text{2.6}}$ \\ \bottomrule
\end{tabular}%
}
\caption{Average agreement ($\times$ 100) in \plms{}' preference on stimuli containing original (\textit{A wug is a \texttt{[CONCEPT]}.}) and alternate framing of novel taxonomic information (\textit{A wug is a type of \texttt{[CONCEPT]}.}).}
\label{tab:agreementtypeof}
\vspace{-1em}
\end{table}


\subsection{How does performance on \compsbase{} vary by property type?}
\begin{figure*}[h]
    \centering
    \includegraphics[width=\textwidth]{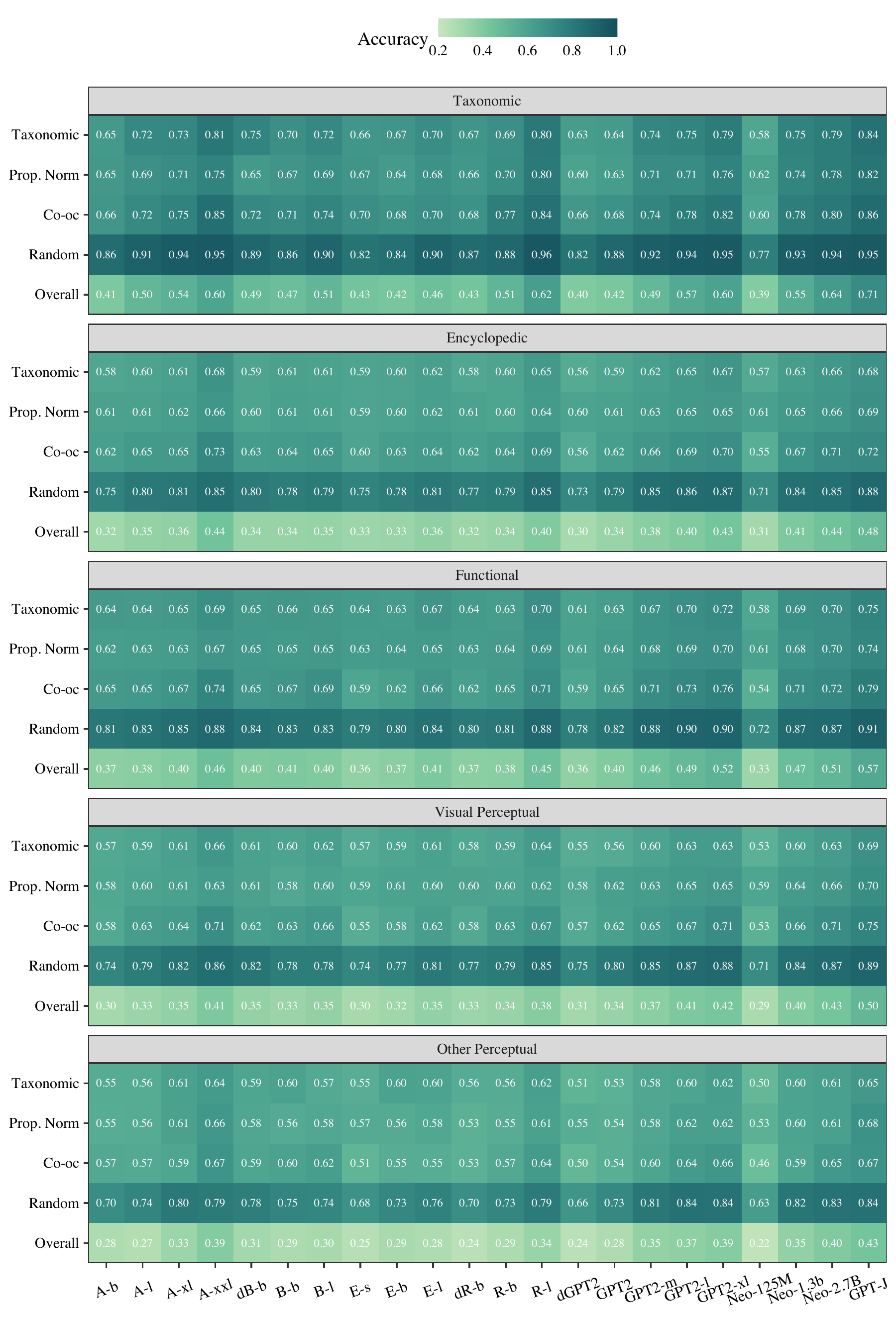}
    \caption{\compsbase{} performance across five property types annotated in CSLB \citep{devereux2014centre}.}
    \label{fig:compsbaseprop}
\end{figure*}

\citet{devereux2014centre} have categorized the properties that we use in our experiments to lie in 5 different categories: (1) \textbf{Taxonomic}, e.g., \textit{is a mammal}, \textit{is a vehicle}, etc.; (2) \textbf{Functional}, e.g., \textit{can keep the body warm}, \textit{is used to hit nails}, etc.; (3) \textbf{Encyclopedic}, e.g., \textit{uses electricity}, \textit{is warm blooded}, etc.; (4) \textbf{Visual Perceptual}, e.g., \textit{has webbed feet}, \textit{has thick fur}, etc.; and (5) \textbf{Other Perceptual}, e.g., \textit{makes grunting sounds} and \textit{is sharp}, etc.
We report results of the 22 \plms{} on the \compsbase{} stimuli across the five different property types, in \Cref{fig:compsbaseprop}.

From \Cref{fig:compsbaseprop}, we observe that \plms{} are substantially stronger in eliciting taxonomic properties of concepts as compared to other types, with highest overall accuracy being 71\%, as compared to 48\% on encyclopedic properties, 50\% on visual perceptual properties, 57\% on functional properties, and 43\% on non-visual perceptual properties. Recall that chance accuracy for the `Overall` scenario is just 6.25\%, so these scores are fairly high.
This corroborates evidence from previous work in analyzing property knowledge of distributional semantic models as well as LM representations to lack perceptual knowledge \citep{lucy-gauthier-2017-distributional, da-kasai-2019-cracking, rubinstein-etal-2015-well, weir2020probing}, likely due to reporting bias \citep{gordon2013reporting, shwartz-choi-2020-neural}. However, different to most of these works, the gap between performance on perceptual properties and non-perceptual properties is small. We conjecture that this could be primarily due to the extension of the CSLB by \citet{misra2022property}, which lead to an increase in coverage of property knowledge for several properties. For instance, the property \textit{has teeth} was mentioned only for 45 out of 67 potential concepts, having been left out for concepts such as \textsc{calf},\footnote{the young one of a cow, and not the muscles in the vertebrate body} \textsc{buffalo}, \textsc{kangaroo}, etc. So it could be the case that previous research has underestimated the extent to which property knowledge is encoded by \plms{} and other distributional semantic models of language. 

\subsection{Does performance on \compsbase{} depend on scale?}
\begin{figure*}[!t]
    \centering
    \includegraphics[width = \linewidth]{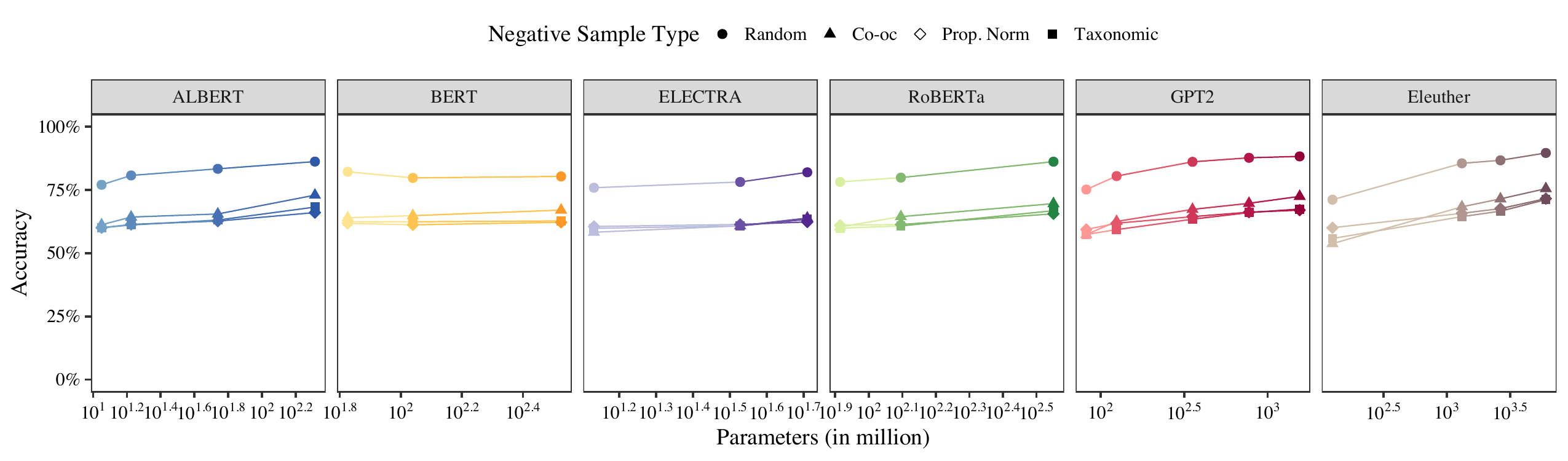}
    \caption{Accuracy vs. parameters across various negative sampling strategies. Models are shaded based on number of parameters.}
    \label{fig:compsbaseparam}
\end{figure*}

We plot the accuracies of \plms{} on \compsbase{} per model family (in order to control for differences in training corpora and tokenization) in \Cref{fig:compsbaseparam}. 
In all families except BERT, we see that accuracy increases with the model size, following standard scaling laws. We notice that distilBERT-base \citep{sanh2019distilbert} is able to outperform even BERT-large on stimuli with `Random' negative samples, suggesting that pruning BERT might sometimes unintentionally improve the model's ability to associate properties and concepts.
We do however caution against interpreting these results as robust conclusion for scaling laws on \compsbase{}. Such an endeavor would require comparing performance of models across multiple checkpoints with varying number of parameters, paired with rigorous statistical inference \citep{sellam2021multiberts, zhang-etal-2021-need}.


\end{document}